# Enhancing an Intelligent Digital Twin with a Self-organized Reconfiguration Management based on Adaptive Process Models

Timo Müller[a,*], Benjamin Lindemann[a], Tobias Jung[a], Nasser Jazdi[a], Michael Weyrich[a]

[a]Institute of Industrial Automation and Software Engineering, University of Stuttgart, Pfaffenwaldring 47, 70550 Stuttgart, Germany

* Corresponding author. Tel.: +49-711-685-67292; fax: +49-711-685-67302. E-mail address: timo.mueller@ias.uni-stuttgart.de

**Abstract**

Shorter product life cycles and increasing individualization of production leads to an increased reconfiguration demand in the domain of industrial automation systems, which will be dominated by cyber-physical production systems in the future. In constantly changing systems, however, not all configuration alternatives of the almost infinite state space are fully understood. Thus, certain configurations can lead to process instability, a reduction in quality or machine failures. Therefore, this paper presents an approach that enhances an intelligent Digital Twin with a self-organized reconfiguration management based on adaptive process models in order to find optimized configurations more comprehensively.

## 1. Introduction

The concept of cyber-physical systems (CPSs) gains more and more importance in industrial automation, as is shown in a countless amount of literature, such as [1–4]. According to [5] the main aspects of such CPSs are, besides their physical components, their connectivity and their abilities for information processing. Production systems consisting of several CPSs are also called cyber-physical production systems (CPPSs) [1].

In addition to the emerging trend of CPPSs in industrial automation, the "increasing volatility in the global and local economies, shortening innovation and product life cycles, as well as a tremendously increasing number of variants, call for production systems, which comply with these changing demands" [6]. These changing requirements lead to objectives for production systems that become increasingly unpredictable in the system design phase. Therefore, the adaption of systems during operation, i.e. reconfiguration, becomes the rule rather than the exception [7, 8].

To address this challenge, an automated reconfiguration management is needed, as derived in [9, 10]. CPPSs offer promising potentials for such an automated reconfiguration management, e.g. through the models of the individual CPPS components which are needed, to identify an existing reconfiguration demand as well as to determine and evaluate alternative configurations. However, existing approaches that aim to realize reconfiguration management only use models created during engineering. A Digital Twin, on the other hand, always has synchronous models of the real system and collects a large amount of process data during the entire life cycle [11]. Therefore, it is obvious to use the Digital Twin also for a reconfiguration management to get much more accurate models, which increase the quality of the reconfiguration.

Models that are even more accurate can be obtained by combining the synchronized models of the Digital Twin with process data being incorporated into the Digital Twin. In this case, it is possible to continuously improve the models by the acquired process data using learning techniques.

Therefore, this paper combines a reconfiguration management with an intelligent Digital Twin, whose models are continuously improved by applying learning techniques on acquired process data in order to further increase the quality of the reconfiguration management and to obtain more reliable and optimal results. Thus, this functionality is added to the capabilities of the intelligent Digital Twin.

The rest of this paper is structured as follows: Chapter 2 gives an overview of work related to reconfiguration management and the intelligent Digital Twin. Chapter 3 explains how adaptive process modeling approaches can be used to continuously improve model accuracy, contains a detailed description of the reconfiguration management methodology, and incorporates both concepts in an intelligent Digital Twin. Chapter 4 provides a conclusion and an outlook on future work.

## 2. Related work

*2.1. Reconfiguration Management*

The topic of reconfiguration covers more than merely the conduction of reconfiguration measures. Thus, the term *reconfiguration management* is specified in [9] and [10] to span the *identification of reconfiguration demand*, the *generation of alternative configurations,* the *evaluation of configurations,* the *selection of a new configuration* and, as an optional extension the *execution of reconfiguration measures.*

The authors of [12] describe an assistance approach for the reconfiguration of CPPSs, for human-robot interactive assembly process. Therefore, they convert the AML file representing the current CPPS structure into a UML data model in order to conduct an attribute mapping between the CPPS capabilities and the production requirements based on defined rules. Whenever a reconfiguration demand is identified, best practice solutions are recommended to assist industrial operators.

In [13] an approach which combines optimization with material flow simulation is utilized to derive optimized reconfiguration scenarios for production systems. Their decision support system contains the four components *optimization* (using CPLEX), *layout generation* (using the CRAFT-algorithm), *simulation* and a *key performance indicators (KPI) dashboard* to enable the comparison of reconfiguration scenarios based on stakeholder-specific KPIs.

An integrated reconfiguration planning tool consisting of a product lifecycle management system and a process simulation for flexible assembly systems is presented in [14]. The authors employ an extended entity-relationship data model, which contains a product-process-resource, a simulation and a production program partial model. The user is assisted in the generation of the simulation model and the simulation execution for different planning alternatives and is provided with evaluation parameters for comparing alternatives.

The framework described in [15] aims to bring self-organizing and self-adaption capabilities to the intelligent shop floor. The approach is based on CPSs and agents, in order to realize a fast allocation of resources in accordance with the production requirements and to reduce disturbances. A gray relational analysis is utilized to find the most appropriate assignment of tasks to machines within the given flexibility corridor of a CPPS during operation.

This excerpt is part of an extensive literature review, which leads to the conclusion that currently no comprehensive reconfiguration management approach with respect to the requirements formulated in [10] is presented by other authors. Furthermore, the reviewed approaches lack the usage of adaptive process models. These models can possess different structures, such as internal and external dynamics [16], or require different amounts of a priori knowledge, such as white-box [17], grey-box [18] and black-box approaches [19]. Based on the existing prerequisites, the selection and usage of appropriate process modeling approaches can improve the knowledge generation based on acquired operating data within Digital Twins [14].

*2.2. Intelligent Digital Twin*

There is a variety of architectures for Digital Twins, [20] has investigated several architectures and presents its own architecture of an intelligent Digital Twin, which combines all determined essential aspects of other architectures. Since this architecture thus draws a rather complete picture of the intelligent Digital Twin, this architecture is taken as the basis for this paper. According to [20], the main components of the intelligent Digital Twin are: a unique ID, synchronous models that can be adapted to the real system at runtime, active process data aquisition, the ability to co-simulate with other Digital Twins, and intelligent algorithms that both enhance the Digital Twin itself and can interact with other intelligent Digital Twins via services. Other components are also mentioned, such as the technical and organizational documentation of the real system. However, these are not relevant for this contribution, therefore they will not be listed here. Further research approaches emphasize the enhancement of intelligent Digital Twins with transfer learning [21].

## 3. Self-organized Reconfiguration Management based on an Intelligent Digital Twin

For the scope of this research, the production cycle of a discrete production system can be divided into two main phases. The first phase represents the reconfiguration and the second phase corresponds to the operation of the production system in the new configuration.

*3.1. Concept for Adaptive Process Modeling*

The cyber-physical production modules (CPPMs) of the CPPS offer services that execute discrete production processes. Between two process executions, the module waits in an idle or standby state for the next execution to be conducted. Hence, the behavior of each module is modeled as a finite state machine with two state types, namely standby and service. This section deals with adaptive process models for production processes that are executed within the service states and that learn during the operating phase. Each of these production processes is designed and evaluated in simulation studies prior to the initial operation of the associated modules. For this purpose, the process engineer usually generates numerical simulation models, which contain the process physics according to the known material and temperature laws. Thus, these models encapsulate the entirety of the existing expert knowledge of the process engineer [22]. At the design time of the module, unknown correlations are not considered in the simulation models. Hence, the basic assumption of all data-driven approaches is that the expert-generated process models do not represent all relations existing in reality, because certain effects and patterns are unknown at design time. The operating phase shall now be used to enrich the existing models with knowledge gained from data and to specify them more accurately by means of a learning process. Different applications can benefit from adaptive models. In particular, an improved decision making in the scope of reconfigurations is realizable. Processes in discrete manufacturing can be characterized by nonlinear autoregressive exogenous models (NARX models) which describe the influence of the applied control variables on the process output [17].

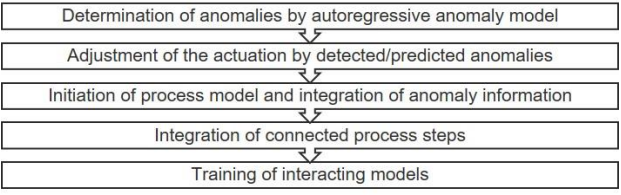

Fig. 1 Steps to generate Adaptive Process Models

The process of generating the model is visualized in Fig. 1 and described in the following. In the case of anomalies that, for instance, affect the actuation system this adaption process can be realized by a model extension regarding the actuating variables. The process model is realized as NARX model and extended by the autoregressive modeling of anomalies (3) being also projected onto a nonlinear time-variant model (4) and linked (2) to the process model (1):

$$\hat{\underline{y}}_n(k) = f_n\left(\underline{y}_n(k-1), \dots, \underline{y}_n(k-n_y), \hat{\underline{u}}_n(k), \dots, \tilde{\underline{u}}_n(k-n_u), k\right) \quad (1)$$

$$\text{with} \quad \hat{\underline{u}}_n(k) = \underline{u}_n(k) + \Delta\hat{\underline{u}}_n(k) \quad (2)$$

$$\text{and} \quad \tilde{\underline{u}}_n(k-\tau) = \underline{u}_n(k-\tau) + \Delta\tilde{\underline{u}}_n(k-\tau) \quad \tau = 1, \dots, n_u \quad (3)$$

$$\text{and} \quad \Delta\hat{\underline{u}}_n(k) = h_n\left(\Delta\tilde{\underline{u}}_n(k-1), \Delta\tilde{\underline{u}}_n(k-2), \dots, \Delta\tilde{\underline{u}}_n(k-n_u), k\right) \quad (4)$$

where the disturbance estimation is represented by $\Delta\hat{\underline{u}}_n(k)$ in (4) and the past observed disturbances $\tilde{\underline{u}}_n(k-\tau)$ are used as arguments and taken into consideration within the horizon $n_u$. The estimation of the refined process output $\hat{\underline{y}}_n(k)$ is based on the past observed external disturbances $\Delta\tilde{\underline{u}}_n(k-1), \dots, \Delta\tilde{\underline{u}}_n(k-n_u)$ and the estimation of the current disturbance $\Delta\hat{\underline{u}}_n(k)$ [23]. This model describes the influence of the detected anomaly patterns on the actuating variables and thus indirectly on the process outputs of a module. To detect and model such unknown effects, and thus to improve the accuracy of the process model, neural networks can be used for $f_n(\cdot)$ and $h_n(\cdot)$ [24]. These neural networks are utilized in the operating phase of the production system and self-adapt $f_n(\cdot)$ and $h_n(\cdot)$ based on acquired data. Thus, previously unknown patterns and relations being relevant for the configuration decision can be integrated into the modeling. With each reconfiguration cycle, more accurate models can be accessed so that a reconfiguration management benefits sustainably and in the long term.

When determining the optimal system configuration, the modeling accuracy of the underlying modules that create the respective system configurations plays a major role. Hence, it is beneficial for the evaluation of a system configuration to include the entire knowledge about existing dependencies between module configurations and to continuously extend this knowledge on the basis of data. The connection between processes is established by the output of the preceding process step, namely the properties of the product, marking an input of the consecutive process step. The modeling approach is extended accordingly:

$$\hat{\underline{y}}_n(k) = f_n\left(\underline{y}_n(k-1), \dots, \underline{y}_n(k-n_y), \hat{\underline{u}}_n(k), \dots, \tilde{\underline{u}}_n(k-n_u), \underline{w}_n(k), k\right) \quad (5)$$

$$\text{with} \quad \underline{w}_n(k) = \underline{y}_{n-1}(k-1) \quad \forall k \in \mathbb{N} \quad (6)$$

This incremental learning process enables the self-organized reconfiguration management to an improved consideration of dependencies between modules when deciding on a new system configuration. For a faster and more robust learning process, a model exchange between different CPPSs with similar or identical modules could take place so that learned anomalies and dependencies can be taken into account when reconfiguring the CPPS under consideration [25].

To map the model outputs $\hat{\underline{y}}_n(k)$ onto three the general evaluation criteria of time, energy and cost, that are needed for the reconfiguration management, the necessary information is extracted and separated. Firstly, equation 5 is written in matrix notation:

$$\hat{\underline{y}}_n(k) = f_n\left(Y_{n,k-1}, \tilde{U}_{n,k}, \underline{w}_n(k), k\right) \quad (7)$$

$$\text{with} \quad Y_{n,k-1} = \left[\underline{y}_n(k-1), \dots, \underline{y}_n(k-n_y)\right] \quad (8)$$

$$\text{and} \quad \tilde{U}_{n,k} = \left[\hat{\underline{u}}_n(k), \dots, \tilde{\underline{u}}_n(k-n_u)\right] \quad (9)$$

where matrix $Y_{n,k-1}$ represents all previous output vectors and matrix $\tilde{U}_{n,k}$ the current and the previous actuation of module $n$. Based on this matrix notation, sub-matrices can be deduced that capture the parts of the process model that are relevant for the criteria time, energy and cost. In the following, the derivation of this parts of the process model is exemplarily described for one criterion and can be easily transferred to all further criteria. The sub-matrix of $Y_{n,k-1}^z$ regarding the general criterion $z$ that contains all relevant relations for that criterion can be described as follows:

$$Y_{n,k-1}^z = Y_{n,k-1}^{IJ} = \left(y_{n,k-1}^{ij}\right)_{i \in \{1,\dots,m_y\}\backslash I^z, \; j \in \{1,\dots,n_y\}\backslash J^z} \quad (10)$$

$$\text{with} \quad I^z \subseteq \{1, \dots, m_y\} \text{ and } J^z \subseteq \{1, \dots, n_y\} \quad (11)$$

$$\text{and} \quad z \in \{time, energy, cost\} \quad (12)$$

where $I^z$ and $J^z$ describe the subsets of the matrix dimensions that are not considered for criterion $z$. The parameters $m_y$ as well as $n_y$ denote the number of process parameters for the regarded service and the time horizon of previous input-output-pairs being relevant for the autoregressive model, respectively. The sub-matrix of actuation $\tilde{U}_{n,k}^z$ that contains all relevant relations for criterion $z$ can be calculated accordingly.

Hence, equation 7 can be reformulated for each criterion of time, energy and cost:

$$\hat{\underline{y}}_n^z(k) = f_n^z\left(Y_{n,k-1}^z, \tilde{U}_{n,k}^z, \underline{w}_n^z(k), k\right) \quad (13)$$

where the part of the process model capturing the relevant relations for $z$ is denoted by $f_n^z(\cdot)$. The model output $\hat{\underline{y}}_n^z(k)$ includes all factors that influence the regarded criterion. These influencing factors have to be transformed to the target criterion and summed up over all conducted process cycles $p$ for the considered module $n$. This results in the following formulation for each target criterion of each module in the service state:

$$f_{n,z,service} = \sum_{k=1}^{p} g_n^z\left(\hat{\underline{y}}_n^z(k)\right) \quad (14)$$

where $g_n^z(\cdot)$ represents the mapping function for all influence factors of a criterion to the criterion itself.

The presented adaptive process models can be used to improve a self-organized reconfiguration management, as the following chapter shows.

*3.2. Self-organized Reconfiguration Management*

The methodology for the self-organized reconfiguration management is depicted in Fig. 2 and evolved from the basic concept introduced in [9]. The concept specifies the conduction of the four steps *identification of reconfiguration demand*, *generation of alternative configurations*, *evaluation of configurations* and *selection of a new configuration*.

Based on this concept, [10] gives a more detailed first concretization of the first two steps. However, the *optimization of production parameters* is not described within the scope of that contribution. Therefore, the further evolved methodology is described, whilst focusing the latter, the evaluation and the selection.

*3.2.1. Identification of reconfiguration demand*

To identify an existing reconfiguration demand, a comparison between the target production and the current configuration of the CPPS is realized utilizing the interface-oriented, formalized process description presented in [26] as a conceptual basis for the modeling. I.e. the functional modeling of the resources capabilities and the production order is realized by a process operator with its input and output state description. On the one hand, this specifies the possible transformations that a production resource offers, and on the other hand, the transformations required for a particular production order.

In order to enable a quick comparison of production requirements and the capabilities of the production system, a CPPS capability model is employed to reveal all possible production sequences [10]. Whenever no possible production sequences can be derived from the CPPS capability model, a reconfiguration demand has been identified and the *generation of alternative configurations* follows.

*3.2.2. Generation of alternative configurations*

The first sub-step of the *generation of alternative configurations* is the *generation of alternatives for production sequences*.

In the beginning, all CPPMs are provided with a description of the desired output product, after which the CPPMs conduct the integrated *generation of alternatives at machine level*. Therefore, each CPPM determines whether it can offer any process operators to reach the desired output product, either in its current or an alternative configuration (at module, i.e. machine level).

Consequently, a new system configuration is created for each of the found process operators. These system configurations comprise the newly found process operator, with its corresponding CPPM configuration, connected to the desired output product of the production order. Thereafter, a new (sub) production order, originating from the corresponding system configuration, is submitted to the CPPMs and the outlined procedure continues, until the defined input product is met or no more suitable process operators can be found. By doing so, each system configuration organizes its own production sequence, resulting in a decentralized, parallelizable approach where a tree consisting of branches represented by alternative system configurations is formed.

The *determination of layout variants of the alternatives* for each of the found possible system configurations is conducted by utilizing a simple brute force approach. Thereby, this sub-step relies on the given layout structure of the CPPS. Thus, the layout of the CPPS is modeled as a graph in which the possible machine locations for the CPPMs are numbered and transport connections between them are represented through the usage of nodes and edges. At this point, it is crucial to determine the effort for the reconfiguration measures at system level $a_{z,reconfiguration}$ that result from the transition into the respective new system configuration, based on the layout information of the current configuration and the reconfiguration efforts of the CPPMs, for each of the criteria $z$ (time, cost and energy). As a result, all different layout variants are represented through a respective system configuration.

The last sub-step is the *optimization of production parameters of production steps*. It aims on minimizing the production efforts $a_{production}$ with regard to the weighted criteria (time, cost and energy) for each alternative system configuration by performing a simulation-based multi-criteria optimization. Therefore, a hierarchical combination of simulation and optimization is employed.

Due to the complexity of the CPPS control logic, the objective function cannot be expressed in a closed form. Consequently, the simulation-based optimization is realized through a category D approach [27], where the hierarchical combination is realized through the integration of the simulation within the optimization.

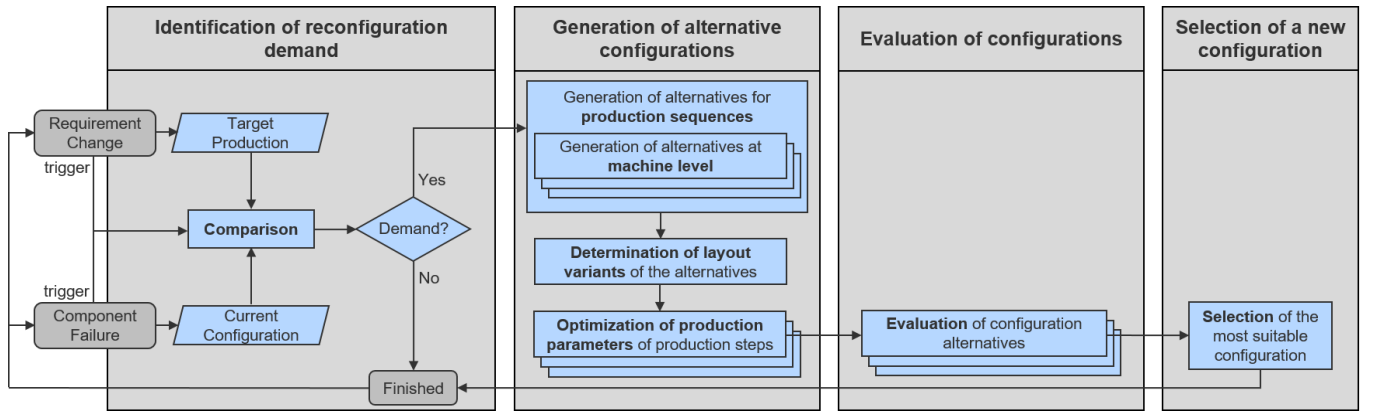

Fig. 2 Reconfiguration Management Methodology

Therefore, the standby effort $f_{n,z,standby}$ with regard to a criterion $z$ is calculated as:

$$f_{n,z,standby} = \sum_{k=1}^{s} g_n^z(c_{n,z,standby}, k) \quad (15)$$

where $c_{n,z,standby}$ originates from the model of the respective module configuration, and is summed up over all occurring standby cycles $s$ for the considered module $n$.

The total effort $f_{n,z,tot}$ of a respective module $n$ with regard to the criteria $z$ can then be determined as:

$$f_{n,z,tot} = f_{n,z,standby} + f_{n,z,service} \quad (16)$$

with $f_{n,z,service}$ from (15).

Consequently, the outputs of the simulation, which represent the resulting production efforts with regard to the respective criteria $z$, can be calculated as:

$$f_z = \sum_{n=1}^{n_{max}} f_{n,z,tot} \quad (17)$$

summed up over all modules $n$. Note that the calculation for the time criterion $f_{time}$ forms an exception and can simply be determined as:

$$f_{time} = f_{1,time,tot} \quad (18)$$

as the sum of the standby and service cycles time of each module (e.g. module 1) already corresponds to the time needed to complete the entire production order. To transform the multi-criteria problem into an optimization problem with only one objective function, and to realize the above mentioned, the following procedure is conducted for each alternative system configuration.

First, a simulation model for the system configuration of the CPPS is built utilizing the modeling concept of state machines to set up the discrete event simulation. Here, the adaptive process models for production processes offered by services are used. Furthermore, to simulate the real production behavior, the applied control logic of the CPPS is depicted within the simulation.

To achieve uniform scaling of the objective functions, each determined system configuration is first optimized separately for each individual objective criterion $z$ (time, cost and energy). From the individual optimization results the reference range between the minimum value $f_{z,min}$ and the maximum value $f_{z,max}$ of a target criterion $z$ can be determined. Note that $f_{z,max}$ corresponds to the highest value of $f_z$ occurring amongst the results of the individual optimization of any criterion except $z$. Thus, the objective functions of the criteria can be normalized and the generalized overall optimization problem then reads as follows:

$$\min F(U) = \sum_z w_z \cdot \frac{f_z(U) - f_{z,min}}{f_{z,max} - f_{z,min}} \quad (19)$$

where $w_z$ is the weight of a respective criterion z, with $\sum_z w_z = 1$ and $U$ is the set of adjustable production parameters for all modules $n$ of a system configuration. It applies:

$$U = [\underline{u}_1(k), ..., \underline{u}_{n_{max}}(k)] \quad (20)$$

The outcome of this step is a set of system configurations with optimized production parameters $U^*$ that furthermore contains the respective optimization result (i.e. the optimized production efforts), which can subsequently be used for a comparison.

*3.2.3. Evaluation of configurations*

For the evaluation of the system configurations, a cost-utility analysis is carried out. Thereby, a utility value $v$ is determined for each system configuration depending on the effort value and criteria weighting.

First, the reconfiguration efforts $a_{z,reconfiguration}$ and the production efforts $a_{z,production}$ (i.e. $f_z(U^*)$ with $U^*$ determined through the optimization) of a system configuration are summed up for each evaluation criterion z:

$$a_{z,tot} = a_{z,reconfiguration} + a_{z,production} \quad (21)$$

Following, for each criterion z the maximum effort value $a_{z,tot,max}$ and the minimum effort value $a_{z,tot,min}$ are determined amongst all possible system configurations. These are assigned to the evaluation value $r_z = 0$ (maximum value) and $r_z = 1$ (minimum value). The evaluation values $r_z$ for each respective criterion z of the remaining system configurations are calculated using the following formula:

$$r_z = \frac{a_{z,tot} - a_{z,tot,max}}{a_{z,tot,min} - a_{z,tot,max}} \quad (22)$$

Thus, for each effort value $a_{z,tot}$ of a system configuration a normalized evaluation value $r_z$ is obtained. Each of a system configurations evaluation value $r_z$ is then weighted by means of the respective criteria weighting $w_z$, and summed up with the results of the other criteria. Thus, the utility value $v$ for each system configuration is calculated as follows:

$$v = \sum_z w_z * r_z \quad (23)$$

*3.2.4. Selection of a new configuration*

To determine the most suitable configuration, all system configurations are compared according to their utility value $v$, with the most suitable system configuration corresponding to the highest value of $v$.

The resulting new system configuration encompasses information at machine and at system level that includes the configuration of each CPPM and its positioning within the CPPS layout as well as optimized production parameters. Consequently, these production parameters can be applied to real production with the new configuration.

*3.3. Mapping to the Intelligent Digital Twin*

Although it would be possible to implement the concepts described in this chapter as stand-alone applications without the help of an intelligent Digital Twin, the latter offers all the prerequisites to support the concepts. Therefore, in this subchapter the presented concepts are mapped to the intelligent Digital Twin.

The models needed for reconfiguration management are a mandatory part of the Digital Twin. In addition, the ability for simulation as well as for co-simulation between several Digital Twins, can be used for the *optimization of production parameters of production steps*. Furthermore, an intelligent Digital Twin automatically collects process data, which is needed for the improvement of the models. Moreover, the intelligent Digital Twin includes intelligent algorithms, the manifestation of which realizes the adaption of the process models as well as the reconfiguration management methodology.

This contribution does not provide a holistic realization of an intelligent Digital Twin as described in detail in [20]. However, several components are discussed and a realization for these essential components of the intelligent Digital Twin is proposed. Thus, a contribution to a possible implementation of the intelligent Digital Twin is made.

## 4. Conclusion and Outlook

In this contribution, a comprehensive reconfiguration management methodology is presented.

- The methodology depicts a decentralized, parallelizable approach to perform an autonomous self-organized reconfiguration management.
- Due to the data-driven approach, precise adaptive process models for each CPPMs module configuration are available.
- The optimization of production parameters takes already learned module dependencies into account.
- Consideration of anomalies leads to more realistic models and therefore better results.
- The effects of identified anomalies and disturbance effects are projected on the three superordinate target criteria, namely time, cost and energy.

In consequence, an improved evaluation of the system configurations resulting from the precise adaptive process models of the module configurations can be conducted. These models are included within the intelligent Digital Twin enabling the reconfiguration management to be incrementally optimized with each operational phase following a reconfiguration.

As future work, a simulation-based multi-objective optimization utilizing a genetic algorithm shall be realized and compared to the proposed approach.